# From the DESK (Dexterous Surgical Skill) to the Battlefield - A Robotics Exploratory Study

**Glebys T. Gonzalez**[1]*, **Upinder Kaur**[1]*, **Masudur Rahma**[1]*, **Vishnunandan Venkatesh**[1], **Natalia Sanchez**[1], **Gregory Hager**[3], **Yexiang Xue**[1], **Richard Voyles**[1], **Juan Wachs**[1,2]

[1]Purdue University, 610 Purdue Mall, West Lafayette, IN 47906

[2]Indiana University School of Medicine, 340 W 10 Street, Indianapolis, IN 46202

[3]Johns Hopkins University, 3400 N. Charles Street, Baltimore, MD 21218.

**Abstract**

*Introduction:* Short response time is critical for future military medical operations in austere settings or remote areas. Such effective patient care at the point of injury can greatly benefit from the integration of semi-autonomous robotic systems. To achieve autonomy, robots would require massive libraries of maneuvers collected with the goal of training machine learning algorithms. While this is attainable in controlled settings, obtaining surgical data in austere settings can be difficult. Hence, in this paper we present the Dexterous Surgical Skill (DESK) database for knowledge transfer between robots. The peg transfer task was selected as it is one of the six main tasks of laparoscopic training. In addition, we provide a machine learning framework to evaluate novel transfer learning methodologies on this database.

*Methods:* A database of surgical gestures was collected for a peg transfer task, composed of 7 atomic maneuvers referred to as surgemes. The collected DESK dataset comprises a set of surgical robotic skills using the four robotic platforms: Taurus II, simulated Taurus II, YuMi and the da Vinci Research Kit, Then, we explored two different learning scenarios: no-transfer and domain-transfer. In the no-transfer scenario, the training and testing data were obtained from the same domain; whereas in the domain-transfer scenario, the training data is a blend of simulated and real robot data which is tested on a real robot.

---

* Authors share equal contribution

*Results:* Using simulation data to train the learning algorithms enhances the performance on the real robot where limited or no real data is available. The transfer model showed an accuracy of 81% for the YuMi robot when the ratio of real-to-simulated data was 22%-78%. For the Taurus II and the da Vinci the model showed an accuracy of 97.5% and 93% respectively, training only with simulation data.

*Conclusions:* Results indicate that simulation can be used to augment training data to enhance the performance of learned models in real scenarios. This shows potential for future use of surgical data from the operating room in deployable surgical robots in remote areas.

**Introduction**

Emergency medical operations and procedures in future Multi-Domain Battlespaces call for technological capabilities that are near-autonomous and can provide rapid patient care at the point of injury. Such closed-loop systems may integrate perception and physical action through teleoperation for accurate detection and treatment of casualties en route care. However, hostile conditions in the battlefield can severely hinder teleoperation. High latency, limited transmission rates, intermittent connectivity, and long-distance communication lead to delays, which are frequent challenges associated with teleoperation in austere settings or an active battlefield.[1,2] To mitigate such challenges, prediction of surgical maneuvers can be an effective strategy to achieve a level of autonomy required for effective surgical performance.[3,4] Prediction models require substantial surgical data previously obtained from procedures similar to those found in the austere setting.[5,6] Nevertheless, the availability of surgical data sources, with procedures performed outside the operating room, is limited as they are difficult to collect. Conversely, simulation offers the opportunity to artificially generate abundant robotic data that largely resembles the target data distribution or domain, thereby facilitating better transfer. Thus, we present a dataset, referred to as Dexterous Surgical Skill (DESK), and a framework for transferring knowledge of surgical steps from simulations to the real domain. Dexterous Surgical Skill includes a diverse set of robots with higher environmental variability than the existing surgical libraries.[7,8]

The DESK dataset is a library of surgical maneuvers called "*surgemes*".[4,9] Surgemes are defined as identifiable gestures that comprise a surgery. These gestures must adhere to the following properties: First, the surgeme must recognizable, even through user variation. Second, surgemes must be modular. This means each gesture represents an individual component of a surgery (i.e. insert needle, push needle, pull suture). [18]

We created a dataset containing visual and kinematic data of a total of 624 transfers, performed by three real robots and a simulated robot (da Vinci Research Kit, ABB Yumi, SRI Taurus and a simulated Taurus). The peg transfer task is one of the 6 key tasks in laparoscopic surgery training. This task was selected for DESK because it is comprised of complex bimanual handling of small objects, which is an essential skill during suturing and debris removal in surgery. In addition to the high number of robots in this library, we purposely added variability to the environmental elements (position and orientation of the pegboard and triangles) to capture the stochasticity of the austere scenarios facilitating better generalization.

Finally, we proposed a baseline architecture for transfer learning from a simulated scenario to real scenarios that remarkably boosts the recognition accuracy in real scenarios. This framework takes the common kinematic features, and then uses a machine learning model to classify the surgemes that are present in a procedure. We obtained an average accuracy of 92% for surgeme classification and an accuracy of up to 97.8% in a transfer learning scenario. In the transfer learning scenario, the models were trained using simulation and real robot data and were tested on the real robot data. Thus, our work has the following contributions: 1) An open dataset for one of the fundamental laparoscopic tasks (peg transfer) with multiple robots and added environmental variability, 2) a baseline for transfer learning of surgeme classification using kinematic data.

**Background**

Autonomous robotics has been proposed as a solution for emergency treatment and rescue in remote areas, especially for time-critical situations or in case of perilous and inaccessible environments.[9,10] To achieve autonomy, massive datasets need to be created for effective training of machine learning algorithms and AI architectures. Two of the most prominently used datasets for recognition and assessment of surgical skills are JIGSAWS[7]

(JHU-ISI Gesture and Skill Assessment Working Set) and MISTIC-SL (Johns Hopkins Minimally Invasive Surgical Training and Innovation Center; Science of Learning Institute). Both datasets were created using the da Vinci Surgical System. These datasets are composed of kinematic data (cartesian positions, orientations, angular velocities and end-effector coordinates), video data of the task, and manual annotations of surgical activity segments and skill.[5,7] The surgical procedures are, in turn, decomposed into atomic surgical units, also known as surgemes, which are used for surgical skill modelling.[11] Such surgemes are the basis to recognize patterns in surgical motions[12] and identify surgical activity.[8] The JIGSAWS dataset is available publicly, however MISTIC-SL is not. They have been used for recognizing surgical activities,[5,6,13] expert demonstration,[14] and surgical trajectory segmentation.[15] However, the main drawback of these datasets is the lack of environmental variability and thereby their lack of universality among different settings. Surgeries in uncontrolled settings exhibit a large range of appearance and kinematic variability.[16] Hence, datasets need to incorporate such variability to facilitate true generalizations when performing pattern matching and prediction, which are key operations for autonomous behavior.

Segmenting and classifying surgical tasks into finite set of maneuvers, called surgemes,[17] aids the pattern identification and skill learning associated with that task. Surgemes can assist in objective evaluation of an operator's skills.[17] Time series data of both JIGSAWS and MISTIC-SL dataset have been segmented and classified using various approaches. Initially, Hidden Markov Models (HMM)[4] were used to evaluate surgical skill levels using surgemes. Tao et al used a combined Markov/semi-Markov Conditional Random Field (MsM-CRF) model for giving useful feedback in surgical training.[18] In recent research, Recurrent Neural Networks have been used for identifying surgical activities.[8,12] However, these approaches use datasets with little variability in the setup (same object positions, setup appearances and type of robot) resulting in a training and testing sets that share the same distribution. Thus, they fail to account for the complexity and randomness that can be encountered in a remote scenario.

Autonomous classification and execution of surgical tasks has been previously attempted with variable success.[19,20] Models derived from such studies have reportedly achieved some form of basic autonomy, when adjusted to specific platforms and environments. Thus, they cannot be directly used to train different robots for different surgical tasks. To achieve autonomous operation, it is desirable to use transfer learning with generalizable datasets.

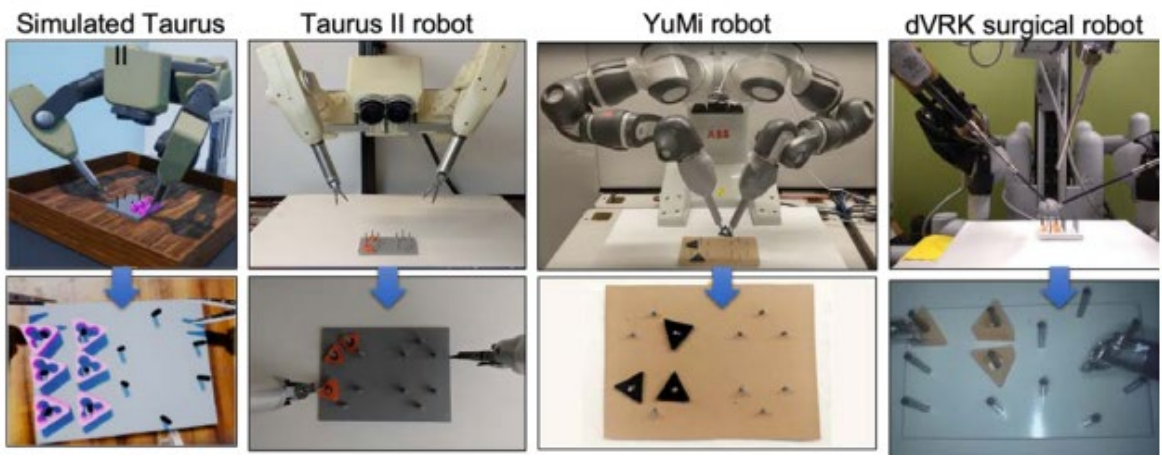

**Figure 1** Data collection setup for the peg transfer task performed by the four robots that comprise the Dextrous Surgical Skill Dataset

Transfer learning, especially in the area of Reinforcement Learning (RL), has been extensively studied in the robotics domain.[21,22] Dimensionality reduction techniques were utilized by Bocsi et al to transfer learning among different robots.[23] In that work, they used a knowledge transfer approach where a bijective mapping was proposed between each dataset and the reduced representations of the datasets, referred to as the lower-dimension manifold. Such methods do not rely on kinematic data of the target domain. Conversely, the method proposed in this paper takes a more contrasting view towards dimensionality reduction. Instead of mapping back the maneuvers to the original space, we implement transfer learning on the reduced space itself by preserving the common features.

**DESK dataset**

Our work presents a test-case for transfer learning between different robots. In addition to this, a database of surgical gestures was collected for the peg transfer task. The collected dataset DESK comprises a set of surgical robotic skills using the four robotic platforms: Taurus II, simulated environment of the Taurus II robot, YuMi, and the da Vinci Research Kit (dVRK) Surgical System, as shown in Figure 1.

**Robot Characteristics**

Taurus II, YuMi and da Vinci surgical robots along with the simulated environments of Taurus II and YuMi were used for creating the dataset. The Taurus II robot is a surgical robot with two 7 degree-of-freedom force feedback manipulators.[16] It was controlled over a stereoscopic display using the Razer Hydra®, as discussed in prior work by the authors.[16] The operator used the two-foot pedals to toggle control between the arms and the camera. In addition, the robot only moved when the clutch pedal was pressed by the operator. The simulated Taurus robot was controlled using the Oculus Rift touch controllers. The setup for the simulator included two pedals to control the robot and to switch between the robot and the camera. For this robot, 16 kinematic variables were recorded including the translation coordinates (x,y,z) with respect to the robot base and the rotation matrix of the wrist (9 values) for both arms.

The YuMi collaborative robot was modified for surgical tasks using 3D printed gripper extensions.[24] The HTC VIVE controllers were used to control the end-effectors of the robot. The RGB video and the depth data was collected using the Realsense® camera. Since, YuMi was designed mainly for industrial tasks with a larger space footprint, the setup in this case was scaled by a factor of two.[16] In this case, 20 kinematic variables were recorded. These variables provide the joint state information and the cartesian (x, y, z translation variables) as well as rotation (with respect to robot's origin) information for the tooltip and the gripper state.

The da Vinci surgical system by Intuitive Surgical (*Sunnyvale, CA, USA*) is widely used in surgical practice throughout the world. In our setup, we used the dVRK controller and two surgical arms to perform the peg transfer task. The cameras in the endoscopic probe were used for recording the videos of the task and the depth data. The *COAG* foot pedal was used by the operator to move the robotic arms and the data was only collected when the pedal was pressed. For this robot, we collected 14 variables for each robot arm along with the timestamp of the execution steps. The angle of aperture of the gripper was recorded as gripper state. The orientation of the gripper was recorded in quaternion format. Unix values were recorded as timestamps.

**Peg Transfer Surgical Training Task**

One of the five tasks used to train residents in laparoscopic surgery, as present in Fundamentals of Laparoscopic Surgery,[25] is the peg transfer task.[26,27] In this task, objects are lifted from one side of a pegboard using one robotic arm, transferred to the other robotic arm mid-air and then subsequently placed in the designated position on the other side of the pegboard. Each subtask of this task requires advanced sensorimotor skills as the clearances between pegs and objects are small. Also, due to the presence of multiple objects in a rather limited space, the maneuverability of the manipulator is also constrained. This task is particularly important as it trains surgeons on bimanual handling of small objects.

The pegboard used for this task has two groups of six poles, as shown in Figure 1 (second row). For the DESK dataset, pegs were picked using one gripper, transferred to the other gripper and then placed at the specified pole on the other side of the pegboard. The position/orientation of the pegboard, the initial and final position of the objects, and the direction of the transfer (objects from right side are transferred to the left side and vice versa) were randomized in order to introduce variability in the dataset as was also done in previous work.[16] Trained non-surgeon operators were used for the data collection.

**Surgemes**

The peg transfer task was subdivided into seven surgical gestures (surgemes). The task starts with *Approach peg* surgeme. Then the gripper is aligned for grasping in the *Align and Grasp* surgeme. Further, the peg is picked up from the pole by the gripper in the *Lift peg* surgeme. Then the grippers are aligned together and moved closer in the *Get Together* surgeme followed by the exchange of peg in the *Exchange* surgeme. The second gripper now moves to the designated pole for the peg in the *Approach Pole* surgeme and places it in the pole in the *Align and place* surgeme, thereby completing the task. The kinematic information, RGB videos and the depth data were annotated with respect to each surgeme. The RGB video files were annotated using a graphical tool developed in-house. The annotation file for each trial contains the name of the surgeme, the start time, and the end time along with the result of the execution, whether the surgeme was a success (Pass) or a failure (Fail). Timestamps of each recording were also stored to allow for synchronization of all the data recording including the depth, kinematic and controller data with the RGB video file.

**Methods**

The approach proposed in this paper uses robot kinematic features to execute transfer learning in surgeme recognition. Since the features used were robot-independent, when performing supervised learning, our method can be applied to different robot domains in transfer learning settings. The applicability to multiple robotic systems is shown in the variability of robots used in our dataset.

**Feature Space for Domain Transfer**

The annotated surgemes within the DESK dataset possess unequal frame lengths. During data processing, every surgeme sequence was re-sampled (via linear interpolation) to fixed number of frames (40 frames) thereby creating a consistent sequential feature vector for every surgeme. Similar to the approach presented in previous work,[16] we condensed the kinematic features collected from each robot to their commonly shared features (7 features each arm) at a base level. Using joint angles would defeat the purpose of robot-independent behavior, because robot arms can reach the same end point with different joint-link configurations. Moreover, the same robot could reach the desired end point with different joint combinations. Therefore, the common kinematic features used were the grippers' position, orientation, and state (open or close). The positional features were measured in the cartesian coordinate system (x,y,z), while the orientation was represented by the angles roll, pitch and yaw. This gives a total of 14 distinct features for each arm (7 each). Apart from these 7 common features, the joint angles and gripper rotation matrices were also recorded. A single 560-dimensional vector per surgeme ($40 \times 7 \times 2$) was created by concatenating 14 features for every frame. Since the task in this case is primarily comprised of linear geometric motions, a Fast Fourier Transform (FFT) was used to map these kinematic features onto a geometric space with respect to each surgeme. Further, supervised learning algorithms were used to recognize and classify surgemes as shown in the architecture in Figure 2. Using a robot-agnostic set of features allows us to leverage information coming from different robotic systems and tasks that differ in scale, position, and appearance (i.e. the pegboard's color and material).

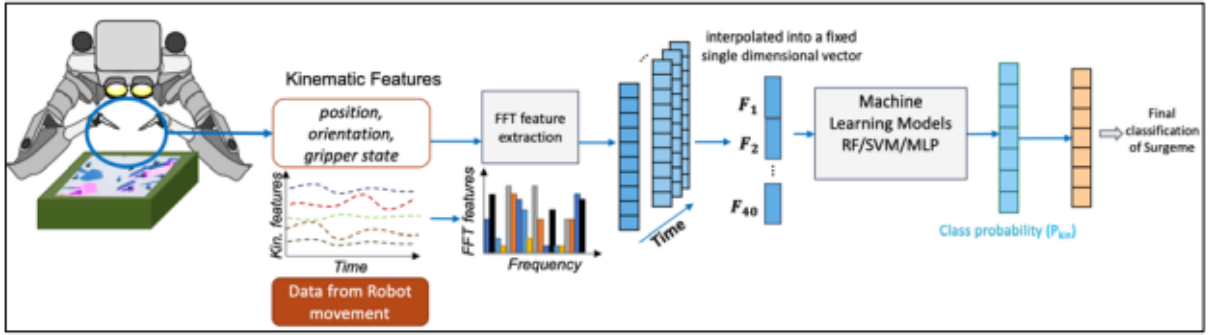

**Figure 2** Architecture overview for Surgeme recognition

**Experimental Setup**

The surgeme classification task experiment was conducted under two scenarios: a) The no-transfer scenario, in which the data used for training and testing comes from the same domain or data distribution (i.e., the same robot). b) The domain transfer scenario, in which the training domain comes from simulation or a mix of simulation and real data, and the testing domain is comprised of the real data.

For the no-transfer setup we compared the performance of the learning models for all the 4 robots in the dataset. Three supervised learning methods were tested for this task: 1) Random Forests (RF), 2) Support Vector Machines (SVM) and 3) A Multi Later Perceptron (MLP). These models received kinematic features of the robot as input and returned as output the class probability of each surgeme. The inputs for these models were the position, orientation and state of both grippers. In the case of the dVRK robot, the joint angles were included, since they improve the classification performance.

Finally, the surgeme classification in each scenario was tested on two levels: Frame-wise and sequence-wise. In the frame-wise setup, a classification was made in video frame, using the kinematic information of that timestep. In the sequence-wise setup, the entire surgeme segment is given to the learning model and the model assigns a surgeme label to the entire sequence.

**Results**

We obtained surgeme classification accuracy results for all the robots using the three supervised learning models (RF, SVM and MLP), as shown in Table 1. These results demonstrate an overall higher accuracy for the sequence-wise setup vs the frame-wise setup, indicating that sequential information is relevant for surgeme identification. In the sequence-wise framework, the beginning and end of the sequence is obtained from the annotations. However, in a live recognition scenario the robotic systems would benefit from a frame-wise classification or from using a learning algorithm designed to take advantage of the sequential information (i.e., Long short-term memory neural networks).

**Table 1** The classification accuracy for each robot in the Dexterous Surgical Skill dataset for the no-transfer scenario. Three classification models were tested: Random forest (RF), Support Vector Machine (SVM) and a Multi-Layer Perceptron (MLP). Results are shown for sequence wise classification and frame-wise classification.

| Robot | Sequence-wise | | | Frame-wise | | |
|---|---|---|---|---|---|---|
| | RF | SVM | MLP | RF | SVM | MLP |
| Taurus II simulator | 88±2 | 87±1 | 78±4 | 86±0 | 58±1 | 73±1 |
| Taurus II robot | 94±2 | 92±1 | 92±2 | 95±0 | 60±0 | 92±1 |
| YuMi robot | 91±1 | 93±1 | 95±1 | 88±1 | 48±1 | 86±1 |
| dVRK surgical robot | 88±3 | 83±3 | 89±2 | 90±0 | 96±0 | 97±0 |

The framework produced models that can accurately classify surgemes. The Taurus II simulator and the real Taurus obtained a maximum accuracy of 88% and 94% respectively, when using Random Forest (RF). For the da Vinci and YuMi robots the maximum accuracy was achieved using MLP, with a 95% accuracy for YuMi and a 97 % accuracy for da Vinci.

Currently, both failed and successful surgemes were assigned the same labels, since the kinematic variables that describe such motions are very similar. Nevertheless, the environment state could look different after a surgeme fails (i.e. a peg was dropped). Future work will focus on leveraging visual features to classify a surgeme as successful or failed.

For the transfer learning scenario, we started by training exclusively with simulation data and testing in the real robot using the architecture described in Figure 2. Then, we increasingly added data from the real scenario to the training set to simulate the effects of limited availability of the real data. We measured the presence of real data in the training model as a ratio between the real and simulated data. When the ratio value is zero, all the data comes from simulation. When the ratio value is 1, the data had a 50%-50% (50/50 = 1) distribution for real-simulated data. Figure 3 shows the classification accuracy of the models for all the real robots when they are trained using real data (orange line) against the performance when the training data is slowly added to the simulation data (blue line).

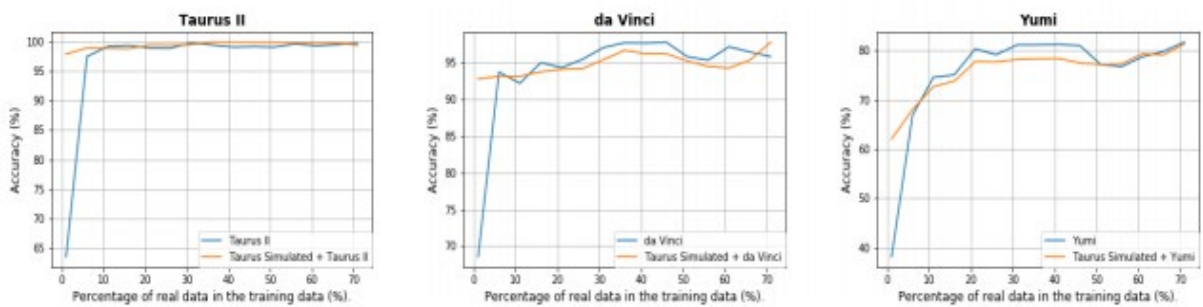

**Figure 3** Performance comparison of training with the real data (no-transfer, shown in orange) vs training with only a percentage of the real data combined with simulation data (transfer learning scenario, shown in blue). The results are shown for the three real robots in the Dexterous Surgical Skill dataset: the Taurus II robot (left) using support vector machines, the YuMi robot (center) using Random Forests (RF) and the dVRK robot (right) using RF.

The Taurus II and da Vinci robots showed a classification accuracy of 97.5% and 93% respectively, even when there were no real examples included in the training set (data ratio=0). Adding a small number of real examples effectively improved the surgeme recognition accuracy. When the ratio of real to simulated data was 15%-85% (ratio=0.18), the classification accuracies went up to 99.7% for the Taurus II and 95.4% for the da Vinci. Therefore, these results show that surgeme classification on real environments can be achieved using a very small amount percentage of real data. The YuMi robot showed a slower convergence, with a ratio of 22%-78% of real to simulated data producing an accuracy of 81%. This accuracy discrepancy is likely due to the YuMi motions. The YuMi robot does not have three degrees of freedom at the gripper, making the orientation changes more abrupt. Thus, the teleoperators would choose a convenient orientation and default to translation motions. In contrast, the gripper's position and orientation changed constantly for the da Vinci, the Taurus II and the simulated Taurus. This inconsistency in the teleoperation resulted in very different FFT features for the YuMi. Thus, it needed more of its own data during training to produce an accuracy over 80%.

Both da Vinci and Taurus II show a faster convergence to a classification accuracy ($\geq$ 95%), needing little to no data in the transfer setup. The FFT features allow to describe the geometric properties (shapes) of the surgemes, while retaining scale and translation invariance. This allows to describe the same trajectory or gripper aperture in robots that have different workspaces. Future work will focus on implementing the same feature engineering strategy on image data to further boost the accuracy of recognition.

**Conclusions**

This work presents a dataset and a machine learning framework to transfer knowledge from different domains in surgical tasks. The dataset includes variability through random changes in the task setup, the orientation of the peg board, and by using different robotic setups with different operators. The dataset has observations from four different robots performing a peg transfer task (624 transfers in total): a simulated Taurus II, a real Taurus II, a YuMi robot and a dVRK surgical robot. The previously available datasets for surgical procedures used a single robotic system for data collection, when skill transfer is required for military robots which have not been used during surgical controlled data collection in the operating room. In the case of remote teleoperation, this constrain is exacerbated as the deployable robot could significantly differ in its kinematic chain configuration and operational space from the surgical robots in an operating room (i.e., da Vinci). Therefore, a surgical task dataset with a higher number of robots and intentionally embedded variability represents a valuable contribution to the research community.

In addition, we presented a baseline framework for surgeme classification in a transfer learning scenario. The transfer model produced an accuracy of 93% and 97.5% for the da Vinci and Taurus II respectively in the extreme case of training with no real data. In addition, the results indicate that using a mix of simulated and real data in the training set can yield higher accuracies. Specifically, training with a ratio of real to simulated data of 22%-78%, YuMi accuracy was boosted from 63% to 81%. With a lower real to simulated ratio (15%-85%), the da Vinci and Taurus II obtained accuracies of 95.4% and 99.7% respectively, as their kinematic data has less discrepancies with the simulated data than the YuMi platform. The results of this work show the potential of using simulated environments to generate data for real, distinct autonomous robots that will be deployed in a remote area.